\documentclass{amsart}

\usepackage{graphicx}

\usepackage{color}		
\usepackage{amssymb,amsmath,amsthm,amscd}
\usepackage{tikz-cd}
\usepackage{xypic}
\usepackage{hyperref}

\newcommand{\catname}[1]{\mathbf{#1}}

\newcommand{\Covf}{\catname{Cov}_{f\ell}}
\newcommand{\Met}{\catname{Met}}
\newcommand{\Meti}{\catname{Met}^\text{inj}}

\newcommand{\Part}{\catname{Part}}
\newcommand{\PerSet}{\catname{PerSet}}
\newcommand{\Percov}{\catname{PerCov}}
\newcommand{\Dend}{\catname{Dendro}}

\newcommand{\Sieve}{\catname{Sieve}}

\newcommand{\R}{\mathbb{R}}
\newcommand{\f}{\mathbf{F}}
\newcommand{\ml}{\mathbf{ML}}
\newcommand{\slc}{\mathbf{SL}}
\newcommand{\lk}{\mathbf{L}^k}
\newcommand{\vlk}{\mathbf{VL}^k}
\newcommand{\elk}{\mathbf{EL}^k}
\newcommand{\gen}{\mathbf{ML}^{\mathcal{T}}}
\newcommand{\complex}[1]{\mathbb{K}\!\left(#1\right)}
\newcommand{\bk}[1]{\mathbf{B}_{#1}}
\newcommand{\bkc}[1]{\mathbf{B}_{#1}^c}

\theoremstyle{plain}
\newtheorem{theorem}{Theorem}

\theoremstyle{definition}
\newtheorem{definition}[theorem]{Definition}

\theoremstyle{definition}
\newtheorem{remark}[theorem]{Remark}

\title[Consistency Constraints for Overlapping Data Clustering]{Consistency Constraints for\\Overlapping Data Clustering}

\author[J. Culbertson]{Jared Culbertson}
\address{Sensors Directorate, Air Force Research Laboratory}
\email{jared.culbertson@us.af.mil}

\author[D.P. Guralnik]{Dan P. Guralnik}
\address{School of Engineering and Applied Science, University of Pennsylvania}
\email{guraldan@seas.upenn.edu}

\author[J. Hansen]{Jakob Hansen}
\address{Applied Mathematics and Computational Science, University of Pennsylvania}
\email{jhansen@sas.upenn.edu}

\author[P.F. Stiller]{Peter F. Stiller}
\address{Department of Mathematics, Texas A\&M University}
\email{stiller@math.tamu.edu}

\date{}	

\begin{document}
\maketitle

\begin{abstract}
We examine overlapping clustering schemes with functorial constraints, in the spirit of Carlsson--M\'{e}moli. This avoids issues arising from the chaining required by partition-based methods. Our principal result shows that any clustering functor is naturally constrained to refine single-linkage clusters and be refined by maximal-linkage clusters. We work in the context of metric spaces with non-expansive maps, which is appropriate for modeling data processing which does not increase information content. 
\end{abstract}

\section{Introduction}
The problem of data clustering has been extensively studied.  Clustering is used in fields as diverse as biology, psychology, machine learning, sociology, image processing, and chemistry, in order to discover hidden structure in data. Among the earliest systematic treatment of clustering theory was that of Jardine and Sibson in 1971 \cite{js-1971}.  Since then, there have been several distinct directions of research in clustering theory, with only modest communication between researchers pursuing different paths. 

The classical work of Jardine and Sibson was followed by other similarly comprehensive works such as Everitt \cite{everitt-2011}. Further theoretical work on these mostly classical methods was also done by Kleinberg \cite{kleinberg} and Carlsson and M\'emoli \cite{cm-2010, cm-2013}.  Work on computing phylogenetic trees inspired a seminal paper by Bandelt and Dress \cite{bandelt-dress} on split decompositions of metrics.  This line of research was continued with investigations into split systems and cut points of injective envelopes of metric spaces.  Representative papers include \cite{dmsw-2013} and \cite{dhm-2001}.  While not explicitly clustering methods, these methods are quite similar in spirit to stratified clustering schemes.  In this category we might also add the classification of injective envelopes of six-point metric spaces by Sturmfels and Yu~\cite{sturmfels-yu}. 

Bandelt and Dress also had a large influence on another field as a result of their work on weak hierarchies \cite{bd-1989, bd-1994}.  This led to work by Diatta, Bertrand, Barth\'{e}lemy, Brucker, and others on indexed set systems (see, e.g., \cite{bbo-2007, bd-2014, diatta-2007}).  An interesting recent development here is the work by Janowitz on ordinal clustering \cite{janowitz}. 

Additional work has been done on topologically-based clustering methods.  This includes the \verb/Mapper/ algorithm by Singh, M\'{e}moli, and Carlsson \cite{smc-2007}, as well as work on persistence based methods 
\cite{cosg-2013} and Reeb graphs \cite{hrpbw-2012}.

Meanwhile, most users of clustering methods default either to a classical linkage-based clustering method (such as single-linkage or complete-linkage) or to more geometrically based methods like $k$-means.  Unfortunately, the wide array of clustering theories has had little impact on the actual practice of clustering.

This paper works to bridge some of these gaps by extending a recent paper of Carlsson and M\'{e}moli \cite{cm-2013}. Their paper introduced the idea of viewing clustering methods as functors from a category of metric spaces to a category of classifying objects giving rise to clusters (e.g. partitions, dendrograms). We will only make use of the very basics of category theory including the notions of categories, morphisms, functors and natural transformations. This abstract language is extremely powerful for not only compactly representing complex information, but also providing a formalism for reasoning about natural operations. For those unfamiliar with these concepts, see either \cite{MacLane:categories} (for a mathematical treatment) or \cite{Barr-Wells:category_theory} (for a computer science perspective). Many desirable properties of a clustering method are subsumed in functoriality when morphisms are properly chosen. One of our principal goals is to extend their theory of functorial clustering schemes to methods that allow overlapping clusters, and in so doing obviate some of the unpleasant effects of chaining that occur for example with single-linkage. Rather than relying on chaining to overcome certain technical problems, we accept overlapping clusters.

\section{Definitions}\label{sec:definitions}
Let $X$ be a set, to be thought of as a set of unlabeled data to be analyzed. In order to make as few assumptions as possible, we only require that $X$ be endowed with a metric, which we will often refer to as $d_X$. However, we do not assume, for example, that $X$ is embeddable in Euclidean space or that $X$ is obtained by sampling from some distribution. Recall that a {\em cover} $\mathcal{C}$ of $X$ is a collection of subsets of $X$ whose union is $X$. A cover $\mathcal{C}$ of $X$ is a {\em partition} of $X$ if $A\cap B=\varnothing$ for any pair of distinct subsets $A,B\in\mathcal{C}$. Also recall that a cover $\mathcal{C}_1$ is said to {\em refine} a cover $\mathcal{C}_2$, if every $U\in\mathcal{C}_1$ is contained in some $V\in\mathcal{C}_2$. 

Traditionally, a clustering method applied to an input dataset is expected to produce a partition of $X$. The work by Kleinberg~\cite{kleinberg} highlights the need for a rigorous treatment of the formal relations between the non-expansive maps among finite metric spaces on the one hand, and refinement relations among partitions produced by distance-based clustering methods on the other. In fact, the main result of {\it loc. cit.} clearly states refinement relations as the obstruction to the ``richness'' axiom (stating that every partition be obtainable as an output of the clustering method for some suitably chosen input, this axiom seems to us as the single least debatable of Kleinberg's axioms). 

Accepting the philosophical position of Carlsson and M\'emoli~\cite{cm-2010} that functoriality of the clustering map is a suitable replacement for the rest of Kleinberg's axioms, our interest in clustering {\em with overlaps} leads us to formulating a restriction on the class of covers of $X$ acceptable as outputs of a distance-based clustering method. However, we prefer to view functoriality as a way of imposing constraints on consistent clustering across datasets, rather than as a set of axioms that must be adhered to. 

Following Jardine and Sibson~\cite{js-1971}, we consider a clustering as encoded by a symmetric and reflexive relation $R$, with clusters being defined as the fibers, $[x]_R:=\{y\,|\,x R y\}$ of the relation. This point of view shows that, in addition to the functoriality constraints already mentioned, a clustering method affording overlaps requires a weakening of the transitivity property (characteristic of partitioning methods). Should transitivity be dropped completely, all that remains is the observation that the fibers of $R$ form a cover of $X$. Still, intuitively, for the purpose of distance-based clustering one feels that three points $x$, $y$, and $z$, which are pairwise ``similar'' to some (measurable) degree need to be regarded as ``jointly'' similar to the same degree. Likewise, this observation should remain valid for larger set of points. This motivates the following definition:
 
\begin{definition}\label{def:flag cover} Let $X$ be a non-empty finite set. A {\em non-nested flag cover} (or simply {\em flag cover}) is a cover $\mathcal{C}$ of $X$ satisfying the following conditions:
\begin{enumerate}
	\item[(1)] If $A,B \in \mathcal{C}$ and $A \subseteq B$, then $A = B$.
	\item[(2)] The abstract simplicial complex $\complex{\mathcal{C}}$ consisting of all the subsets of elements of $\mathcal{C}$ is a flag complex.
\end{enumerate} 
We denote the set of flag covers of $X$ by $\Covf(X)$.\qed
\end{definition}
Note that $\mathcal{C}$ is the collection of maximal simplices of $\complex{\mathcal{C}}$, with a simplex spanning $S\subseteq X$ if and only if $S$ is contained in some element of the cover $\mathcal{C}$. Thus, $\complex{\mathcal{C}}$ is flag if, for every $S\subseteq X$, $S$ spans a simplex in $\complex{\mathcal{C}}$ whenever every pair of distinct points $x,y\in S$ is contained in an element of $\mathcal{C}$.
In particular, every partition of $X$ is a flag cover of $X$.

Finally, note that any cover $\mathcal U$ of $X$ can be ``upgraded'' to a non-nested flag cover $\mathcal U^{\ast}$, commonly referred to as the {\em flagification of $\mathcal{U}$}, in a minimal way, where $\mathcal U$ will refine $\mathcal U^{\ast}$ and $\mathcal U^{\ast}$ will refine any other flag cover which $\mathcal U$ refines. This may be done by iteratively adjoining to $\mathcal{U}$ any clusters mandated by the flag condition, and then removing all the non-maximal ones.

Perhaps the most common notion in the clustering literature (see, e.g., ~\cite{js-1971,diatta-2007}) related to flag complexes is that of {\em maximally linked sets}. We recall:
\begin{definition}\label{def:maximally linked} Let $X$ be a set and let $R$ be a symmetric, reflexive relation $R\subset X\times X$. A subset $S\subseteq X$ is {\em maximally linked with respect to $R$} if (1) $x R y$ for all $x,y\in S$, and (2) $S$ is not properly contained in any subset of $X$ satisfying (1).\qed
\end{definition}
Clearly, picking $\mathcal{C}$ to be the set of all maximally linked subsets of $X$ with respect to $R$ results in a flag cover of $X$. One of the most studied constructions of this form is the {\em Vietoris--Rips complex}, arising from a metric space $(X,d_X)$ as $\complex{\mathcal{C}}$ upon setting $x_1Rx_2\Leftrightarrow d_X(x_1,x_2)\leq\delta$, for some $\delta \geq 0$.

\begin{definition}\label{def:persistent cover} A {\em persistent cover} on $X$ is a function $\theta_X\colon \R_{\geq 0} \to \Covf(X)$ such that
\begin{enumerate}
\item[(1)] If $t_1 \leq t_2$ then $\theta_X (t_1)$ refines $\theta_X (t_2)$.
\item[(2)] For any $t$, there is an $\varepsilon>0$ with $\theta_X (t^\prime )=\theta_X (t)$ for all $t^\prime \in[t,t+\varepsilon )$.
\end{enumerate}
If we also have (3) below, then we call $\theta_X$ a {\em sieve} on $X$: 
\begin{enumerate}
\item[(3)] There exists $t \in \R_{\geq 0}$ such that $\theta_X (t)$ is the trivial cover $\{X\}$.\qed
\end{enumerate}
\end{definition}

\noindent Persistent covers and sieves are a direct generalization of Carlsson and M\'emoli's {\em persistent sets} and {\em dendograms}, which satisfy the same conditions, but have the set of partitions of $X$ as codomain. 
They may also be seen as a sort of strictly isotone indexed set system as in \cite{bertrand-2000}, where the index of each set $A \in \bigcup_{t \in \mathbb{R}_{\geq 0}} \theta_X (t)$ is given by the infimum of the values of $t$ such that $A \in \theta_X (t)$.

We now consider the category $\Met$, which has finite metric spaces as objects and non-expansive mappings as morphisms.  That is, a map of sets $f\colon X\to Y$ is a morphism $(X,d_X )\to(Y,d_Y )$ in $\Met$ if for any 
$x,x' \in X$, $d_Y(f(x),f(x')) \leq d_X(x,x')$.  This is the same as saying $f^* (d_Y )\leq d_X$, where $f^* (d_Y )$ is the metric on $X$ given by $f^* (d_Y )(x_1 ,x_2 )=d_Y (f(x_1 ),f(x_2 ))$.  Note that any morphism $f\colon (X, d_X) \to (Y, d_Y)$ factors through $(X, f^{\ast}(d_Y))$. We abuse terminology somewhat by allowing zero distances between points in our finite metric spaces. In this way $(X, f^{\ast}(d_Y))$ is a valid object in $\Met$ even when $f\colon X \to Y$ is not injective. 

We want to take objects in $\Met$ and convert them into (collections of) clusters in various ways. The category $\Covf$ is the category of ordered pairs $(X,\mathcal{C})$, where $X$ is a set and $\mathcal{C}$ is a flag cover of $X$. A morphism between $(X,\mathcal{C})$ and $(Y,\mathcal{D})$ is a map of sets $f\colon X \to Y$ such that $\mathcal{C}$ is a refinement of $f^{-1}(\mathcal{D})$. These are called {\em consistent maps}. Note that $f^{-1}(\mathcal{D})$ need not be a flag cover of $X$, though it becomes one upon removal of its non-maximal elements.

The category $\Part$ of partitions is a subcategory of $\Covf$, where only coverings that are also partitions are allowed. We define $\Sieve$ as the category of pairs $(X,\theta_X )$, where $\theta_X \colon \mathbb{R}_{\geq 0} \to \Covf(X)$ is a sieve on $X$. The morphisms in $\Sieve$ are an extension of the morphisms of $\Covf$; that is, a set map $f\colon X \to Y$ is a morphism of sieves $(X,\theta_X) \to (Y,\theta_Y)$ if for every $t \in \R_{\geq 0}$, $\theta_X(t)$ refines $f^{-1}(\theta_Y(t))$. Note that this means that we have a family of functors from $\Sieve$ to $\Covf$, by restricting to a particular value of the parameter $t$. For convenience, we summarize these categories in Figure~\ref{fig-table}. 

\begin{figure}[h]
\centering
\begin{tabular}{lll}
{\bf Category} & {\bf Objects} & {\bf Morphisms} \\
$\Met$ & Finite metric spaces $(X,d_X)$ & Non-expansive maps\\
$\Meti$ & Finite metric spaces $(X,d_X)$ & Non-expansive injections\\
$\Part$ & $(X,\mathcal{P}_X)$, $\mathcal{P}_X$ a partition of $X$ & Consistent maps \\
$\Covf$ & $(X,\mathcal{C}_X)$, $\mathcal{C}_X$ a flag cover of $X$ & Consistent maps  \\
$\PerSet$ & $(X,\Phi_X)$, $\Phi_X$ a persistent set on $X$  & Consistent maps \\
$\Dend$ & $(X, \Phi_X)$, $\Phi_X$ a dendrogram on $X$ & Consistent maps\\
$\Percov$ & $(X, \theta_X)$, $\theta_X$ a persistent cover of $X$ & Consistent maps\\ 
$\Sieve$ & $(X,\theta_X)$, $\theta_X$ a sieve on $X$ & Consistent maps \\
\end{tabular}
\caption{Summary description of relevant categories. A consistent map from $(X,\mathcal{C}_X)$ to $(Y,\mathcal{C}_Y)$ is a set function $f\colon X \to Y$ such that for every $A \in \mathcal{C}_X$, there exists some $B \in \mathcal{C}_Y$ such that $A \subseteq f^{-1}(B)$. A consistent map from $(X,\theta_X)$ to $(Y,\theta_Y)$ is a set function $f$ such that for every $t \in \R_{\geq 0}$, $f$ is a consistent map from $(X, \theta_X(t))$ to $(Y, \theta_Y(t))$.}\label{fig-table}
\end{figure}

\section{Flat Clustering} The primary development of this paper focuses on clustering methods that work at a fixed scale, giving clusters of similar data points either as blocks of a partition or sets in a covering. In the next section, we will briefly describe how these methods can be extended to hierarchical versions. 

\subsection{Functors on Met}

We consider a {\em flat} or {\em nonhierarchical (overlapping) clustering} to be a covariant functor $\f$ from $\Met$ to $\Covf$, which restricts to the identity on the underlying set, i.e., $\f(X, d_X)$ takes the form $(X, \mathcal C)$ where $\mathcal C$ is a flag cover of $X$. We will refer to such $\f$ as {\em clustering functors}.  A reasonable first question is whether there are any interesting such functors, and the following definition provides a useful way of constructing many examples. 

\begin{definition}\label{def:gen} Let $\mathcal{T}$ be a set of finite metric spaces. Given a metric space $(X,d_X)$, define a relation $R$ on $X$ with $x R y$ if and only if there exists a morphism $t$ from some $T\in \mathcal{T}$ into $(X,d_X)$ satisfying $x,y \in \text{Im}(t)$. 
Let $\gen(X,d_X)$ be the covering of $X$ by maximally linked subsets under $R$. We refer to $\gen$ as the \emph{clustering functor generated by $\mathcal{T}$}.\qedhere
\end{definition}
\begin{remark}\label{rem:gen} Clearly, the relation $R$ above is preserved under any morphism. By this we mean that if we have a morphism $f\colon X \to Y$ in $\Met$ and $xRy$, so that there is a morphism $t\colon T \to X$ for some $T \in \mathcal T$ with $x$ and $y$ in the image of $t$, then the composition $f \circ t\colon T \to Y$ yields $f(x)Rf(y)$. This verifies that $\gen$ is, indeed, functorial for $\Met$.
\end{remark}

A wide range of clustering functors are obtainable in this way. We begin with the standard single-linkage clustering scheme. Given a parameter $\delta$, we can construct the Vietoris--Rips complex from any metric space $X$ by adding an edge between two points whenever the distance between them is at most $\delta$. Define 
$\mathcal{C}_{\slc}$ as the partition of $X$ given by the connected components of the Vietoris--Rips complex of $X$, and define $\slc_\delta\colon \Met \to \Covf$ by $(X,d_X) \mapsto (X,\mathcal{C}_{\slc})$. Carlsson and M\'{e}moli, in 
\cite{cm-2013}, have shown that $\slc_\delta$ is functorial when viewed as a map to $\Part$. Since $\Covf$ contains $\Part$, the mapping $\slc_\delta$ is also functorial with $\Covf$ as target category. Alternatively, it is easy to see that $\slc_\delta$ is generated by the collection of spaces $\Lambda_\delta^k=\{0,\ldots,k\}$ endowed with the metric $d(i,j)=\delta|i-j|$, with $k$ ranging over the positive integers (see Definition~\ref{def:gen}).

We now define maximal-linkage clustering in a similar fashion. Given $\delta$, again construct the Vietoris--Rips complex of $X$. We then take $\mathcal{C}_{\ml}$ to be the set of maximal simplices of this complex. We define $\ml_\delta$ as the map taking $(X,d_X)$ to $(X,\mathcal{C}_{\ml})$. Alternatively, $\ml_\delta$ is $\gen$ for $\mathcal{T}=\{\Lambda_\delta^1\}$. 
\begin{theorem}
$\ml_\delta$ is a surjective functor from $\Met$ to $\Covf$. 
\begin{proof}
The image under $\ml_\delta$ of a morphism $f$ in $\Met$ should be the morphism in $\Covf$ given by the same set function, if it is indeed a morphism in $\Covf$. Thus, as long as $\ml_\delta$ maps morphisms to morphisms, it will respect composition. 
In the following diagram, we need to show that $\ml_\delta(f)$ is a morphism in $\Covf$ given that $f$ is a morphism in $\Met$.
\[
\begin{tikzpicture}
\node (x) at (0,0) {$(X,d_X)$};
\node (y) at (4,0) {$(Y,d_Y)$};
\node (xc) at (0,-2) {$(X,\mathcal{C}_X)$};
\node (yc) at (4,-2) {$(Y,\mathcal{C}_Y)$};

\draw[->, left] (x) to node {$\ml_\delta$} (xc);
\draw[->, above] (x) to node {$f$} (y);
\draw[->, below] (xc) to node {$\ml_\delta(f)$} (yc);
\draw[->, right] (y) to node {$\ml_\delta$} (yc);
\end{tikzpicture}
\]

Define a reflexive symmetric relation $D_X$ on $X$ with $x D_X x'$ if $d_X(x,x') \leq \delta$; similarly, let $y D_Y y'$ if $d_Y(y,y') \leq \delta$. Then for any morphism $f$ in $\Met$, $x D_X x' \Rightarrow f(x)D_Y f(x')$. 
Under a mild abuse of notation, this means that $D_X \subseteq f^{-1}(D_Y)$. Note that the sets in $\mathcal{C}_X$ are the maximal linked sets under $D_X$. Since $f^{-1}(D_Y)$ contains $D_X$, every maximal linked set under $D_X$ is
contained in a maximal linked set under $f^{-1}(D_Y)$. Hence $\mathcal{C}_X$ refines $f^{-1}(\mathcal{C}_Y)$.

An alternative proof of this fact is to note that the sets in $\mathcal C_X$ are the maximal linked sets, i.e. if $U$ is one such set then $d_X (x_1 ,x_2 )\leq\delta$ for every $x_1 \in U$ and $x_2 \in U$, and for every 
$\tilde{x}\in X,\tilde{x}\notin U$ there is an $x\in U$ with $d_X (x,\tilde{x})>\delta$.  It follows that $d_Y (f(x_1 ),f(x_2 )) \leq \delta$ so that all the points in $f(U)$ are within $\delta$ of each other.  They therefore lie in 
some maximal linked set $V$ in $\mathcal C_Y$.  It follows that $U\subset f^{-1}(V)$ and that $\mathcal C_X$ refines $f^{-1}(\mathcal C_Y )$.

To see that $\ml_\delta$ is surjective, note that the cover $\mathcal{C}$ implicitly defines a simplicial complex on $X$ by taking the sets in the cover as maximal simplices. Because $\mathcal{C}$ is a flag cover, this complex is 
flag, uniquely determined by its $1$-skeleton. We can therefore metrize the $1$-skeleton of this complex by setting every edge length to $\delta$, and setting the distance between any two disconnected points to be $2\delta$. Then the 
distance between two points in the complex is less than or equal to $\delta$ if and only if they are in the same simplex. This implies that the maximal simplices are exactly the maximal linked sets under this metric. Thus every 
flag cover arises from some metric on $X$ under the $\ml_\delta$ map, so that this map is surjective.
\end{proof}
\end{theorem}

The concept in the preceding proof of defining a symmetric reflexive relation on $X$ and then taking its maximal linked sets is an important one. We may reformulate $\slc_\delta$ in terms of a relation by letting $x \sim_\delta y$ for two elements $x,y$ of $X$ 
if there is a positive integer $k$ and a sequence of points $x = x_0, x_1, \ldots, x_{k-1}, x_k = y$ such that $d(x_{i-1},x_i) \leq \delta$ for all $1 \leq i \leq k$. This is just a more explicit way of saying that there is a non-expansive map $\Lambda_{\delta}^k \to X$ containing $x,y$ in its image (see the definition of $\slc_\delta$ above).  In this case the relation is an equivalence relation and the single-linkage clusters are simply the equivalence classes of $\sim_\delta$. Similarly, $\ml_\delta$ consists of the maximal linked sets of the relation $M$ where $x M y$ if $d(x,y) \leq \delta$. 

This suggests the possibility of expanding the relation $M$ to include more pairs but not to the extent of the relation $\sim_\delta$. One potential method is to fix a positive integer $k$ and define a relation $R^k_\delta$ as before, such that $x R^k_\delta y$ if there 
exists a sequence of $k+1$ points (not necessarily all distinct) $x = x_0, x_1, \ldots x_{k-1}, x_k = y$ such that $d(x_{i-1},x_i) \leq \delta$ for $1 \leq i \leq k$. In other words, we can get from $x$ to $y$ in $k$ steps of size at most $\delta$. We denote the resultant map from $\Met\to\Covf$, given by taking 
maximal linked sets of $R_\delta^k$, as $\lk_\delta$, and call it {\em $k$-linkage clustering}. Observe $\lk_\delta$ is generated by $\mathcal{T}=\{\Lambda_\delta^k\}$. In particular, $\lk_\delta$ is a functor by Remark~\ref{rem:gen}.

Of course, additional relations are possible. For example, we could also define $R^K_\delta$ where we can take as many steps as we like provided that the sum of the lengths are no more than $K$. Alternatively, we could combine this with $R^k_\delta$ to obtain the relation $R^{k, K}_\delta$ where we require that $\sum_{i=0}^{k-1} d(x_i, x_{i+1}) \leq K$. 

An immediate consequence of the definition of $\lk_\delta$ is that for any metric space $(X,d_X)$ and any threshold value $\delta \geq 0$, there exists $k \in \mathbb{N}$ such that $\lk_\delta$ is equivalent to $\slc_\delta$ on $(X,d_X)$.
This may be summarized as saying that $\slc_\delta = \mathbf{L}_\delta^\infty$. Similarly, $\ml_\delta = \mathbf{L}_\delta^1$.  The functor $\mathbf{L}^{2}_{\delta}$ is also known as ``Cech clustering at scale $\delta$'' or sometimes ``at $2\delta$''.

Now let $\f\colon\Met\to\Covf$ be any (flat/non-hierarchical) clustering functor.  We consider the two-point metric space $\Lambda_{\varepsilon}^{1}$ with distance $\varepsilon\geq0$ between the two points.  Note that if $\varepsilon'\geq\varepsilon$ then there is a non-expansive mapping (morphism in $\Met$): 
\[
	\Lambda_{\varepsilon'}^{1}\to \Lambda_{\varepsilon}^{1}.
\]
Thus if $\f(\Lambda_{\varepsilon}^{1})$ consists of two single point clusters then so does $\f(\Lambda_{\varepsilon'}^{1})$.  On the other hand, if $\f(\Lambda_{\varepsilon}^{1})$ is a single cluster, then so is 
$\f(\Lambda_{\varepsilon''}^{1})$ for any $\varepsilon''\leq\varepsilon$.

We call $\f$ trivial if $\f(\Lambda_{\varepsilon}^{1})$ is two single point clusters for all $\varepsilon\geq0$ or if $\f(\Lambda_{\varepsilon}^{1})$ is a single two point cluster for all $\varepsilon\geq0$.  One can easily show 
that in the former case $\f(X)$ is the cover by singletons for all $(X,d_X )$ in $\Met$ and in the latter case $\f(X)$ is just the cover consisting of $X$ itself for all $(X,d_X )$.  (Keep in mind that the cover $\f(X)$ is a flag
cover.)

Thus if $\f$ is non-trivial there exists a number $\delta_\f \geq0$ such that $\f(\Lambda_{\varepsilon}^{1})$ is a single two point cluster if $\varepsilon<\delta_\f$ and two singleton clusters if $\varepsilon>\delta_\f$.  The question 
of what happens when $\varepsilon=\delta_\f$ is a minor annoyance, and we will assume $\f(\Lambda_{\delta_\f}^{1})$ is a single two point cluster.  The other case can be handled with some minor changes to our discussion.

\begin{definition}
 Given a non-trivial clustering functor $\f$, we call $\delta_\f$ the {\em clustering parameter} for $\f$.\qed
\end{definition}

Note that if $\f$ has clustering parameter $\delta_\f$ and $(X,d_X )$ is any metric space with $x_1 ,x_2 \in X$ and $d_X (x_1 ,x_2 )\leq\delta_\f$ then $x_1$ and $x_2$ lie in a common cluster (set of the cover) of $\f(X,d_X )$.

\begin{theorem}
Suppose $\f$ is a non-trivial clustering functor $\Met \to \Covf$ with clustering parameter $\delta_\f$. Then for any input space $(X,d_X)$, the output of $\f$ refines the output of $\slc_{\delta_\f}$ and is refined by the output of $\ml_{\delta_\f}$.
\begin{proof}
Suppose $x,y \in X$ such that $d_X(x,y) \leq \delta_\f$. Then there exists a morphism $\Lambda_{\delta_\f}^1\to(X,d_X)$ with image $\{x,y\}$. By the hypothesis, $\f$ merges $\Lambda_{\delta_\f}^1$ into one cluster, so there must be some 
cluster $A$ in $\f(X,d_X)$ such that $x,y \in A$. Here we are using the functoriality of $\f$, which means we have a morphism 
\[
	\f(\Lambda_{\delta_\f}^1) \to \f(X, d_X) = (X, \mathcal C_X)
\]
in $\Covf$, and our single two point cluster must refine the pullback of $\mathcal C_X$ to $\Lambda_{\delta_\f}^1$. Since $\f(X,d_X)$ is flag, if $B$ is a maximal linkage component of $(X,d_X)$ at scale $\delta_\f$ then $B$ is contained in an element of $\f(X,d_X)$.

Now suppose $x,y$ are elements of $X$ such that $x$ and $y$ are in separate components of $\slc_{\delta_\f}(X, d_X)$. Then there exists a morphism $(X, d_X) \to \Lambda_\varepsilon^1$ for some $\varepsilon > \delta_\f$ which sends $x$ and $y$ to different 
points in $\Lambda_{\varepsilon}^{1}$.  But this implies that $x$ and $y$ can never be in the same cluster in $\f(X, d_X)$.
\end{proof}
\end{theorem}

Note that this does not imply that the clusters in $\f(X, d_X)$ are unions of Rips clusters (i.e., $\ml_{\delta_\f}(X, d_X)$), which is false in general. 

\subsection{Functors on $\Meti$}
Carlsson and M\'emoli, after proving the uniqueness of $\slc$ as a functor $\Met \to \Part$, considered an expanded class of functors, those from $\Meti \to \Part$. In this section we consider some other clustering schemes in this context.

A number of overlapping clustering schemes have been suggested in the literature. Jardine and Sibson \cite{js-1971} proposed two ``type B'' methods that restricted the size of the overlap between clusters. We consider these two methods, along with two similar methods based on $k$-vertex and $k$-edge connectivity. The $\bk{k}$ clustering method is designed to prohibit overlaps of cardinality greater than or equal to $k$. One way to obtain it is by taking the maximally linked clusters for a given level $\delta$, and repeatedly merging any two clusters that overlap in $k$ or more points. Alternately, one may construct the threshold graph for a given $\delta$, and then repeatedly add edges implied by the following property: if $a$ and $b$ are vertices, and there exists a complete subgraph $S$ of size $k$ such that both $a$ and $b$ are adjacent to every vertex in $S$, then $a$ and $b$ are adjacent. Then the $\bk{k}$ clusters are the maximal cliques of this graph. This requirement is relaxed in the coarser method $\bkc{k}$, in which $S$ need not be a complete subgraph, or even connected at all, i.e., $a$ and $b$ are each adjacent to a subset $S$ of $k$ points. Note that neither $\bk{k}$ nor $\bkc{k}$ are functorial for $\Met$ (see figure~\ref{fig-vertexce} below).

\begin{figure}[t]
\centering{\includegraphics{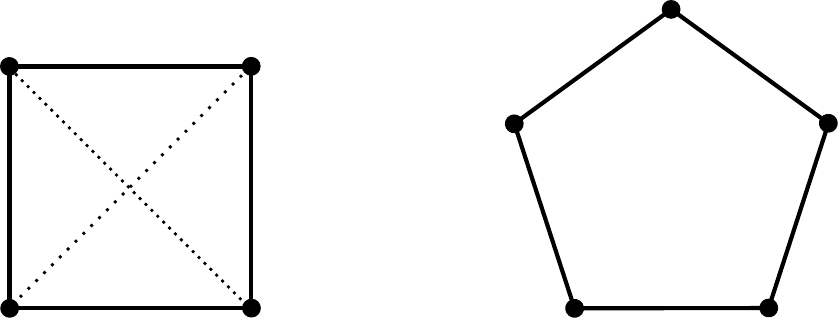}}
\caption{Both these graphs are $2$-connected, and if maximal would be clusters under $\mathbf{VL}^2$. However, the left graph would also be a cluster under $\bkc{k}$, as the dotted edges are induced. Neither graph is a cluster under $\bk{k}$, since the maximal complete subgraphs are simply the edges, which overlap in sets of cardinality $1$.}\label{fig-linkcomparison}
\end{figure}

We define the $\vlk$ clustering methods as follows: Given a metric space $(X,d_X)$ and $\delta \geq 0$, we construct the graph $G$ with with vertices equal to the set $X$, where there is an edge between $x$ and $y$ if and only if $d_X(x,y) \leq \delta$. We call this graph the {\em $\delta$-threshold graph} for $(X, d_X)$. Then for any integer $k \geq 1$ construct the covering of $X$ given by maximal $k$-vertex-connected subgraphs of $G$. We denote this clustering method $\vlk$. Note that $\slc_\delta = \mathbf{VL}_\delta^1$. Further inspection shows that $\lim_{k\to\infty} \vlk_\delta = \ml_\delta$. All three of these methods $\bk{k},\bkc{k}$ and $\vlk$ are distinct, as Figure~\ref{fig-linkcomparison} shows. 

The use of $k$-vertex connectivity in defining the $\vlk$ clustering methods leads naturally to the idea of using $k$-edge-connectivity to separate clusters. Note that the maximal $k$-edge-connected subgraphs always form a partition of the vertices, unlike the maximal $k$-vertex-connected subgraphs. We will call this clustering method $\elk$. As with vertex connectivity, we also have that $\mathbf{EL}^1_\delta = \slc_\delta$ and $\lim_{k \to \infty} \elk_\delta = \ml_\delta$ for all $\delta \geq 0$. In general, however, $\elk$ and $\vlk$ will produce different results.

\begin{figure}[h]
\def\svgwidth{\columnwidth}
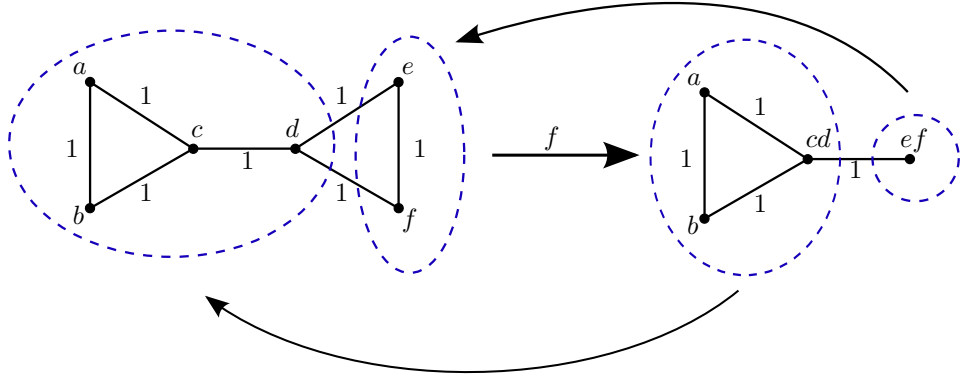
\caption{The induced clusters from the non-expansive map $f$ do not refine the original $2$-edge-connected clusters, showing that $\mathbf{EL}^2_1$ is not functorial on $\Met$.} \label{fig-EL}
\end{figure}

It is easy to see that each of these clustering methods fails to be functorial on $\Met$ for finite $k > 1$ and any $\delta \geq 0$. Consider the two spaces in Figure~\ref{fig-vertexce}. For $\delta = 1$, $X$ is grouped into a single cluster under the three overlap-restricting methods. However, the non-expansive mapping $f$ takes $X$ onto a metric space that has two clusters. The lack of functoriality stems from the restriction on numbers of overlapping points. Morphisms in $\Met$ may collapse several points into one, thus splitting a $k$-vertex-linked subgraph. Similarly, the two spaces in Figure~\ref{fig-EL} show that $\elk_\delta$ ($k > 1$) is not functorial on $\Met$, with the problem again arising from the fact that multiple points can be collapsed into a single point. This motivates the consideration of the category $\Meti$ as in \cite{cm-2013}, which restricts morphisms to injective non-expansive maps. 

\begin{figure}[h]
\def\svgwidth{\columnwidth}
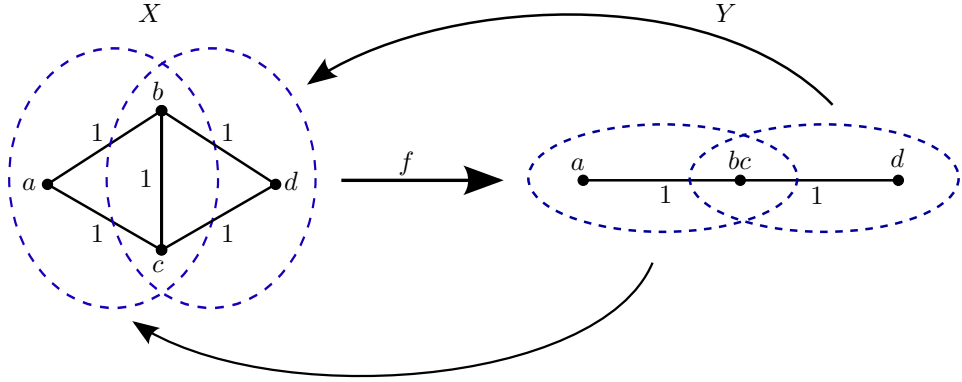
\caption{For these metric spaces, all three overlapping clustering methods give the same result. The preimage of the $\bk{2}$, $\bkc{2}$, or $\mathbf{VL}^2_1$ clusters in $Y$ is finer than the clustering in $X$, so none of these methods is functorial over $\Met$. }\label{fig-vertexce}
\end{figure}

\begin{theorem}
The mappings $\bk{k},\bkc{k},\vlk_\delta$ and  $\elk_\delta$ from $\Meti \to \Covf$ are functorial for each $k \geq 1$ and $\delta \geq 0$.
\begin{proof}
Let $f\colon (X,d_X) \to (Y,d_Y)$ be a morphism in $\Meti$. If $G_X$ and $G_Y$ are the $\delta$-threshold graphs associated to $(X, d_X)$ and $(Y, d_Y)$, then $f$ induces a graph homomorphism $G_X \to G_Y$ since $f$ is injective and non-expansive. Thus $f$ preserves both edge and vertex connectedness, and the maximal edge or vertex-connected subsets of $G_Y$ contain the images of the maximal connected subsets of $G_X$. In other words, $\elk_\delta(X, d_X)$ refines $f^{-1}\left(\elk_\delta(Y, d_Y)\right)$ and $\vlk_\delta(X, d_X)$ refines $f^{-1}\left(\vlk_\delta(X, d_X)\right)$, as desired.

The proof for $\bk{k},\bkc{k}$ follows directly from results of Jardine and Sibson~\cite{js-1968,js-1971}. 
\end{proof}
\end{theorem}

\begin{theorem}
For every $\delta \geq 0$, there is a sequence of natural transformations 
\[
\ml_\delta = \mathbf{VL}^\infty_\delta \to \cdots \to \vlk_\delta \to \cdots \to \mathbf{VL}_\delta^1 = \slc_\delta
\] in the category of functors $\Meti \to \Covf$.
\begin{proof}
Note that if $k \leq \ell$, the clustering given by $\mathbf{VL}^\ell_\delta$ always refines the clustering given by $\vlk_\delta$. Then the identity maps from $\ml(X, d_X)$ to $\slc(X, d_X)$ are morphisms in $\Covf$ for any $X$. 
\end{proof}
\end{theorem}

The biconnected components of a graph can be computed in linear time; given a division of a graph into maximal $k$-connected subgraphs, these can be divided into $k+1$-connected subgraphs by finding all $k$-element vertex cuts. This can be done in polynomial time for each $k$, so the $k$-vertex connected components of a graph can be enumerated in polynomial time. (For more information see \cite{matula-78}). Constructing the adjacency graph for a given metric requires quadratic time in the number of points, so the $\vlk_\delta$ clustering schemes can be calculated in polynomial time for any fixed $k$. However, the maximal clique problem is NP-complete, so no polynomial time algorithm is known to compute $\ml_\delta$ in general.

Note that the $\mathbf{EL}_\delta$ method is excisive in the sense of Carlsson--M\'emoli~\cite{cm-2013} for each $\delta \geq 0$, so by Theorem $6.2$ of {\it loc. cit.}, it is representable by a set of test metric spaces whose injections into $X$ determine the clusters. However, it may be more efficiently calculated using one of several fast algorithms for finding maximal $k$-edge-connected subgraphs, such as those in \cite{zlylcl-12}, \cite{cyqlll-13}, and \cite{aiy-13}.

\section{Hierarchical Clustering}
All of the parameterized flat clustering schemes we have considered generalize naturally to hierarchical clustering methods which we call {\em sieving functors} $\f\colon\Met \to \Sieve$.

\begin{theorem}
Suppose $\delta \leq \delta'$. Then for any $k$ (including $\infty$) there is a natural transformation $\lk_\delta \to \lk_{\delta'}$.
\begin{proof}
The theorem follows easily from the fact that for any $X$, the clustering given by $\lk_\delta(X)$ refines that given by $\lk_{\delta'}(X)$ whenever $\delta \leq \delta'$.
\end{proof}
\end{theorem}

\begin{theorem}
\label{thm-functoriality}
Suppose $\f_t$ is a family of functors from $\Met$ to $\Covf$ indexed by nonnegative real numbers $t$ such that whenever $t \leq t'$, there is a natural transformation $\f_t \to \f_{t'}$. Then the map 
$\f\colon \Met \to \Sieve$ given by $(X,d_X) \mapsto (X,\theta_X)$, with $\theta(t) = \f_t(X, d_X)$ is a functor.
\end{theorem}

The proof of Theorem~\ref{thm-functoriality} follows easily from the definition of a sieve, and we call a functor of this type a {\em sieving functor}.  The two previous theorems then give us a family of hierarchical clustering schemes, $\{\ml,\ldots, \lk,\ldots,\slc\}$. Note, however, that there are many more functorial hierarchical clustering schemes. A broader theoretical treatment will be given in a forthcoming paper, ``Functorial Clustering via Projections,'' where we work with sets having more general dissimilarity measures and provide a characterization of stable sieving functors. 

\section{Acknowledgements}The authors are grateful for the financial support of the Air Force Office of Scientific Research under the LRIR 12RY02COR, LRIR 15RYCOR153, and MURI FA9550-10-1-0567 grants. 

\bibliography{metric_spaces}
\bibliographystyle{plain}

\end{document}